# DUET: Detection Utilizing Enhancement for Text in Scanned or Captured Documents

Eun-Soo Jung*, HyeongGwan Son*, Kyusam Oh, Yongkeun Yun, Soonhwan Kwon, Min Soo Kim
Technology Research
Samsung SDS
Seoul, Republic of Korea
{es2018.jung, hk86.son, q3.oh, yongkeun.yun, soonhwan.kwon, minsoo07.kim}@samsung.com

*Abstract*—We present a novel deep neural model for text detection in document images. For robust text detection in noisy scanned documents, the advantages of multi-task learning are adopted by adding an auxiliary task of text enhancement. Namely, our proposed model is designed to perform noise reduction and text region enhancement as well as text detection. Moreover, we enrich the training data for the model with synthesized document images that are fully labeled for text detection and enhancement, thus overcome the insufficiency of labeled document image data. For the effective exploitation of the synthetic and real data, the training process is separated in two phases. The first phase is training only synthetic data in a fully-supervised manner. Then real data with only detection labels are added in the second phase. The enhancement task for the real data is weakly-supervised with information from their detection labels. Our methods are demonstrated in a real document dataset with performances exceeding those of other text detection methods. Moreover, ablations are conducted and the results confirm the effectiveness of the synthetic data, auxiliary task, and weak-supervision. Whereas the existing text detection studies mostly focus on the text in scenes, our proposed method is optimized to the applications for the text in scanned documents.

## I. INTRODUCTION

The amount of scanned or captured documents have been increased rapidly with the spread of robotic process automation and digital transformation. However, effective and fast methods for understanding text in the document images are not sufficiently investigated. Especially, text detection for scanned documents has not been studied compared to the detection of text in the wild.

Detecting texts from natural images have been remarkably advanced together with image processing technologies led by deep learning methods. The vision and machine learning community have demonstrated promising results in detecting texts of various conditions and scenes [1]-[14]. Most of the leading or highlighted detection approaches are optimized for text in the wild, thus their performances were also validated with scene text databases. Scene text databases [15]-[19] include images such as street signs, signboards, logos, and car plates. Text in these images has different features from those in documents. Orientations, sizes, and shapes of text in the wild are comparatively more diverse. In contrast, document images usually contain more words aligned in lines with fewer variations in font, size, color, and spacing. There are likely to be more words in a document than in a scene image, whereas the words in documents tend to be in similar conditions and arranged form. Despite these differences between the text in documents and scenes, text detection methodologies for document images are less focused and are mostly adopted from those for scene text images.

In this paper, we propose a text detection method for scanned or captured documents, named as DUET for *detection utilizing enhancement for text*. In this method, we adopt a concept of multi-task learning that combines a segmentation-based detection with an auxiliary task of text enhancement. An ideally enhanced document is a binarized image with clear text without noise. Namely, our proposed network has two branches for text detection and enhancement and learns the two tasks in parallel. Multi-task learning has been used across various applications [20]-[25] and demonstrated with outstanding results. We expect that learning the text enhancement task, such as noise reduction and blur restoration, is beneficial to robust text detection for noisy document images. Moreover, we can additionally achieve the enhanced document images.

In the field of text recognition for documents, one of the difficulties is the insufficiency of labeled data. Annotating each word or character with its location requires considerable labor cost since documents are likely to contain a large amount of text. Fortunately, however, texts in documents have fewer variations in background sceneries, fonts, shapes and orientations, thus we can synthesize the data relatively easier than scene text data. Therefore, we train our model using the synthetic data together with real document images to overcome the shortage of labeled data and increase robustness in text detection.

Whereas the labels of enhanced text masks are easily paired with the synthetic data, the text masks from real document images are relatively harder to achieve. Therefore, we trained our networks in two steps: 1) training with only synthetic data in a fully-supervised manner and 2) training with both synthetic and real data in a weakly-supervised manner. Weak-supervision methods have been used for image segmentation tasks [26], [27], and they demonstrated that class labels or bounding box information could be useful when exact segmentation ground truths are not affordable. In our weakly-supervising phase, the quality of text enhancing is estimated with text box labels and text detection scores from the enhanced outputs.

---

\* Equal contributions.

In summary, we propose a novel text detection method specialized for document images. 1) We introduce text enhancing subtask to extract text features effectively. 2) We generate synthetic document images in order to increase the volume of training data, and 3) exploit weak-supervision for the enhancement of real data. As a result, our method outperforms previous works and its validity is supported with ablation tests.

## II. RELATED WORKS

### A. Scene Text Detection

Network models and learning techniques from object detection and segmentation have been adopted in text detection. The most successful networks for object detection were trained with box regression approaches [24], [25], [28], [29] and their potentials for text detection have been demonstrated with impressive results. Connectionist Text Proposal Network (CTPN) [3] is a joint model of convolutional neural network (CNN) and recurrent neural network (RNN) that proposes text regions and sequential connections. TextBoxes [30], which was inspired by single shot multibox detector (SSD) [28], modified its kernels and anchor boxes for text shapes. A different SSD-based method SegLink [31] detected text segments and links and weaved them into a word or a text line. To detect text in multi- and arbitrary-oriented shapes, several approaches were suggested, such as direct regressions [7], rotation of convolutional filters for rotation-sensitive features [32], and rotation region proposal with rotation region-of-interest (RoI) pooling [9]. Recurrent Neural Networks (RNNs) were used for refining arbitrary text regions in [33]. FOTS [34], an end-to-end text detection and recognition method, introduced RoIRotate to share convolutional features between detection and recognition with an FPN-inspired backbone.

Variations in shapes and directions could be controlled more efficiently with segmentation-based detection methods [35]-[37] that determine text regions in pixel-level. Fully convolutional neural network (FCN) based models are widely used to predict text blocks or segments with text score maps from the previous studies [2], [8], [10], [12].

Regarding a word or text line as a text instance is more general, where the majority of text databases are labeled in word- or line-level, not in character-level. However, Character Region Awareness for Text detection (CRAFT) [13] adopted weak-supervision to train score maps with characters and affinity information and demonstrated text detection in both word- and character-level. Recently, CharNet [14] that conducts both text detection and character recognition in a weakly-supervised manner was also proposed.

Most studies for text detection have focused on resolving difficulties in scene text. Moreover, scene text data [15]-[19] with labels are much richer than document images [38] in quantity and diversity. Consequently, a majority of deep learning models were trained and tested with scene text databases. Commercialized optical character recognition services have adopted techniques of scene text recognitions and demonstrated their performances also with document images. To the best of our knowledge, deep learning models for text detection in document images have not been explored adequately.

### B. Document Image Processing

Learning-based image processing methodologies [39]-[42] have replaced and outperformed hand-designed filters or feature extractors to enhance or restore images. Understanding forms or contents of degraded document images is a challenging task, where overcoming degrading factors such as noise, blur, background patterns, or distortion is essential.

Studies have demonstrated deblurring document images with deep neural networks, such as CNNs [43] and skip-connected deep convolutional autoencoders [44]. Deep learning methods were also adopted for binarizing document images and separated pixels for text and background [45], [46]. DeepErase [47] proposed neural-based preprocessor to erase ink artifacts from text images.

Training a series of scale- and iteration-wise filters was suggested [48] to capture domain-specific properties of document images. Category-specific generative adversarial networks (GAN) were designed for the restoration of face and text images [49], [50]. With 3D geometrical information, networks can train to unwarp geometrically distorted document images [51], [52]. A study [53] conducted text detection together with document binarization with U-net [35] based architecture and demonstrated the benefits of multi-task learning. However, because of the insufficiency of document image databases, especially ground truths for binarizing or restoration, quantitative performances or comparisons of document image processing methods are limited.

## III. METHODOLOGY

### A. Data Synthesize

We synthesized scanned document images to train the proposed network. Parts of sentences crawled from news articles or arbitrary sequences of characters were randomly selected and placed line by line on empty background images in 320 by 320 pixels. Overall 92 characters are included in the data: upper and lower cases of English alphabets, digits, and special characters. Among 32 special characters on QWERTY keyboard layout, 30 characters are included except grave accent (`) and vertical bar (|). The background images are distinctive with various features, such as textures, watermarks, stains, and colors. The sizes, weights, and styles of the text fonts were also randomly selected for each line. Then horizontal or vertical lines were added to represent texts with underlines, tables, or noise from printers. Generated samples were augmented in various ways: blurring, adding various types of noise, reducing resolution. Examples of synthesized images are depicted in Fig. 1a.

Text masks as ground truths for enhancement and word boxes as ground truths for detection were also generated together with the document images. The generation method of binary text masks is same as that of the document images without any background and noise. The binary text masks can be considered as enhanced text images where faded or blurred parts of texts are restored and noises are removed as in Fig. 1b. A word box includes a top left and bottom right corner coordinates of each word. A sequence of characters between spaces is considered as a word regardless of their character types (alphabets, digits, or special characters).

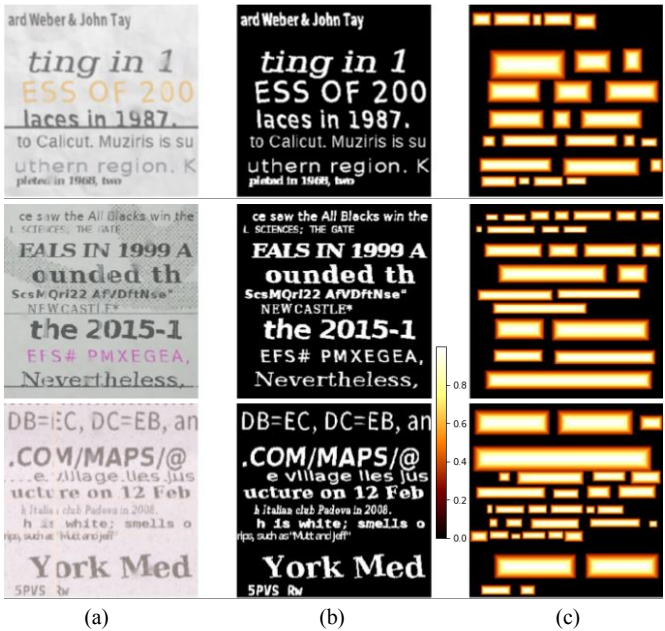

Fig. 1. Examples of synthesized (a) document images, (b) corresponding enhanced text masks, (c) and word score maps.

### B. Score Map Generation

A text score map is generated according to the word boxes for each document image. A character-level score map generation with 2D Gaussian is adopted in [13]. However, this approach is inappropriate to represent score maps for words, which consist of various numbers of characters. The left and rightmost parts of extremely long words can be missed with the simple 2D Gaussian scoring method as depicted in Fig. 2a and b. Therefore, we modified the Gaussian map generation to a rectangular-shaped map as Fig. 2c. A square-shaped score map fit to a shorter side of a word box (usually height) is generated by merging (take the minimum value for each pixel) 1D Gaussian distributions along the x- and y-axis. Then the map is split in the middle and pushed to the edges of the longer side of the word box, and the remaining middle part of the box is filled with values as the middle line where the square map is split. An example of the square-shaped score map and its transformation into a rectangular-shaped word box is depicted in Fig. 2d. More examples of score maps are depicted in Fig. 1c.

### C. Network Architecture

The principal concept of DUET is taking advantage of multi-task learning by adding an auxiliary task, which is text enhancement; we adopt a fully convolutional encoder-decoder network with an additional decoder branch as Fig. 3a. An output of the detection branch is a detection score map, and that of the segmentation branch is an enhanced text mask for an input document image. As Feature Pyramid Network (FPN) [54] based network architectures with ResNeXt [55] have been utilized for various detection tasks and recorded state-of-the-art performances, an FPN-based network is adopted as the backbone of our proposed network.

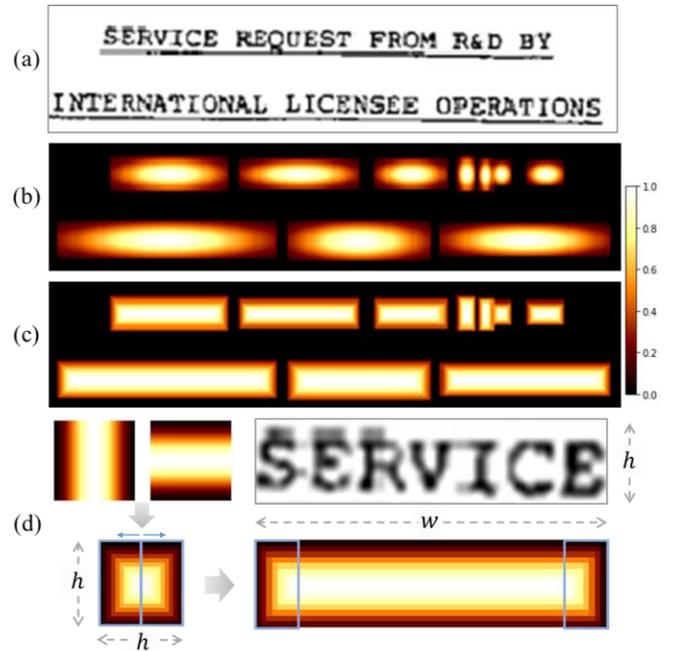

Fig. 2. An example of (a) text image with words of various lengths and (b) its simple 2D Gaussian score map and (c) modified rectangular-shaped score map. The first and last characters of longer words such as 'INTERNATIONAL' and 'OPERATIONS' are not fully covered as other characters in the middle with the simple 2D Gaussian. (d) An example of a word box generation when the height ($h$) of the word box is shorter than its width ($w$).

### D. Training

We define a multi-task loss for text detection and enhancing as:

$$L = \lambda L_D + (1-\lambda) L_E, \quad (1)$$

where $L_D$ is a loss between predicted detection score and ground truth detection score maps ($GT_D$) and $L_E$ is a loss between predicted and ground truth text enhancement masks ($GT_E$), and the two terms are balanced with the parameter $\lambda$ ($\in [0, 1]$). To achieve the loss from each branch, ground truths for both detection and enhancement are required. The ground truths of enhancement and detection maps for the synthesized text images that we have introduced are generated together with the images, however, the real document databases with text detection labels are not paired with binarized text masks. Therefore, we have divided the training process into two phases: *phase-1* with only synthetic data and *phase-2* with both synthetic and real data.

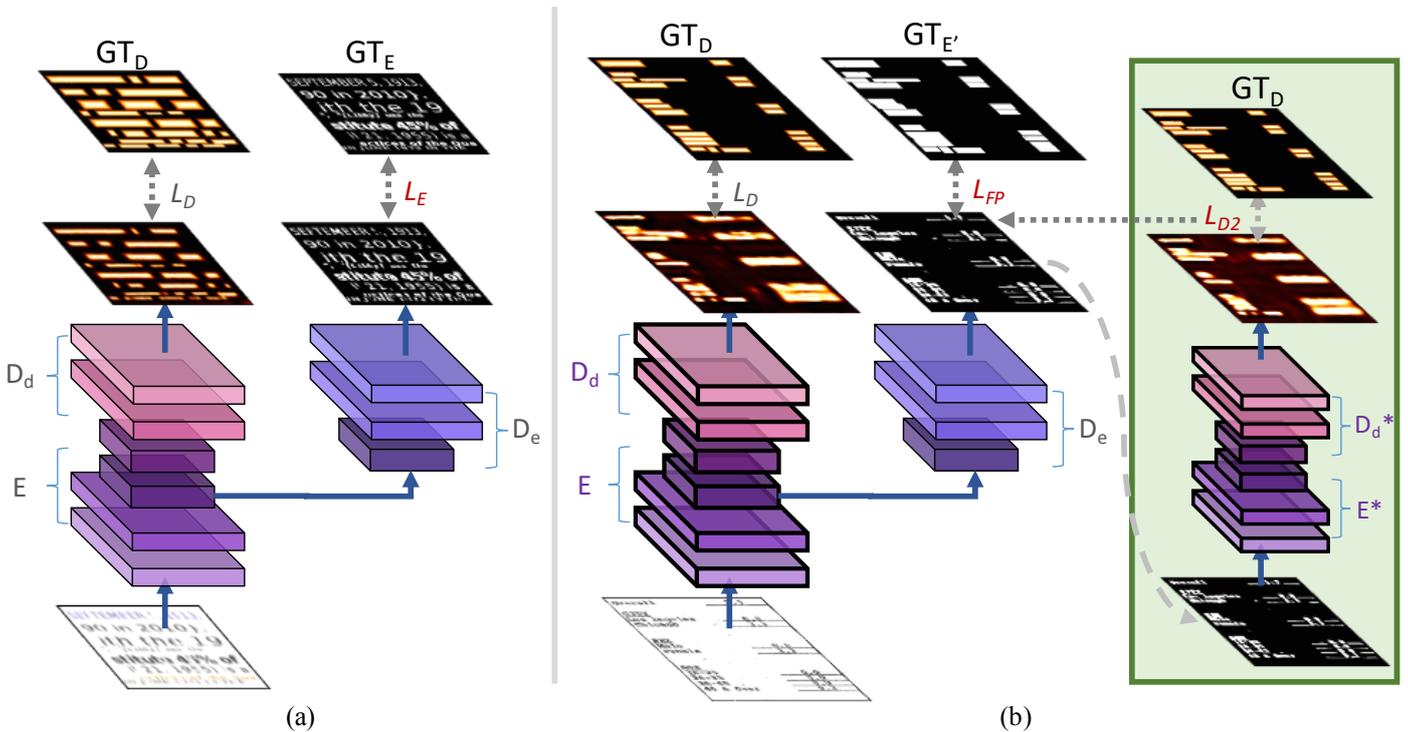

Fig. 3. Overall architecture of DUET and its learning strategy in (a) training *phase-1* and (b) training *phase-2* with real data. Enhancement of a real image is weakly-supervised with its binarized ground truth of detection map and the detector part of trained interim model; the encoder (E) and detection decoder ($D_d$) of the interim model are duplicated every batch in phase-2 as E* and $D_d$*, perform text detection for the enhanced output, and provide detection loss as the quality of the enhancement. Details of network backbone, such as the number of layers or skip connections, are simplified.

*1) Training phase-1*

In training phase-1, each training sample has its ground truth for both text detection and enhancement. Interaction over union (IoU) and $L_2$ losses are adopted for the text enhancement and detection task, respectively. The proposed network learns to separate pixels of characters from background and to detect areas for text in a fully-supervised manner in this phase. Overall network architecture is depicted in Fig. 3a.

*2) Training phase-2*

Real document data are added in training phase-2, which have ground truths for detection but not for enhancement. Therefore, the detection loss is achieved as in training phase-1 but weak-supervision is adopted to train the enhancement task. We implement two ways to evaluate enhancement results with the ground truths for detection.

First, a detection score map does not provide an exact text mask but does provide background area with no text; the regions with zero scores in detection ground truths should be also labeled as background for enhancement. Therefore, false positive losses can be roughly estimated with binarized detection ground truths ($GT_{E'}$). With synthetic images as inputs, the false positive losses are achieved with ground truth text masks.

Also, the detector of the model trained up to date can be exploited for the evaluation of enhancement. The detection part of trained interim DUET is duplicated every batch and this detector conducts text detection of the enhanced text mask outputs. As an ideal enhanced text mask can be regarded as a text image without any noise or blur, lower detection errors are expected with better enhancement results. The detection loss is achieved with $L_2$ loss as that for the detection branch.

Consequently, instead of the IoU segmentation loss in (1), enhancement loss ($L_E$) is replaced with false positive loss ($L_{FP}$), detection loss ($L_{D2}$), and their balancing parameter ($\lambda_2$) in the training phase-2:

$$L_E = \lambda_2 L_{D2} + (1 - \lambda_2) L_{FP} . \quad (2)$$

Training in phase-2 with real data is represented in Fig. 3b.

## IV. EXPERIMENT

*A. Dataset*

**Form Understanding in Noisy Scanned Documents (FUNSD) dataset** [38] contains overall 199 annotated document images: 120 images for train, 29 for validation, and 50 for test. The images have various widths ranged in 754 to 863 with fixed heights of 1000 pixels. For each document image, locations and text labels are annotated in word-level. Location of a text box is represented with two coordinates: top-left and bottom right coordinates. Also, words are grouped into semantic entities and labeled for entity type (*i.e.* question, answer, or header) and relation with other entities. In total, there are 31,485 words and 9707 semantic entities.

## B. Implementation Details

Including augmented ones, we generate 60k synthetic document images for training phase-1 and add 360 real images for training phase-2. To increase the volume of the real data, each real image is augmented three times by resizing (to 2000 pixels in height) and random cropping (to 1600 by 800, to cover various shapes of inputs such as scanned receipts). During training phase-1, the synthetic data are used without resizing. During training phase-2, the input size is set at 1600 by 800 with three channels as augmented real images, and the synthetic images are randomly placed on a 1600 by 800 empty canvas. Also, the ratio of the synthetic and real data in train phase-2 is 5:1, which is empirically determined (1:5, 1:1, 100:1 are also tested).

FPN [54] network with ResNeXt [55] of depth 101 layers is the backbone of DUET, and architectures of the detecting and enhancing branches are identical. The balancing parameter between detection and enhancement loss is set at 0.5, and the parameter for weakly-supervised enhancement losses (the false positive loss and detection loss) in training phase-2 is also set at 0.5. The warm restart technique proposed in [56] is adopted.

## C. Post-processing

Text boxes are selected from a score map output with the post-processing method introduced in [13]. Non-Maximum Suppression (NMS) method is unnecessary for our approach. However, DUET can provide two text detection outputs: the detection of an original input and that of the enhanced result. Additionally, we ensemble the two results and reduce false detections; match text boxes detected from the original to those from the enhanced inputs with IoU, and discard small boxes (height shorter than 10 pixels) from the original with no match (IoU < 0.01).

## D. Results

We compare the text detection performances of DUET to those of the leading research or systems for text detection. Precision, recall, and F-scores set at IoU = 0.5 are presented in TABLE I. The results of Tesseract [5], EAST [12], Google Vision API (https://cloud.google.com/vision/docs/pdf), and Faster R-CNN [24] are provided in [38]. Also, we tested CRAFT [13] and CharNet [14].

DUET was trained with relatively small training data, without any scene text dataset, but it outperforms other methods with the highest recall and F-score. Examples of the detection results are visualized in Fig. 4a together with the corresponding enhanced output in Fig. 5a. Text in documents of noisy background patterns is successfully detected. The high recall rates and figures of text boxes indicate that DUET rarely skips characters and even detects hand-written ones, which were not included in the synthetic dataset. However, there are falsely detected boxes of incorrectly merged words or miss alignments.

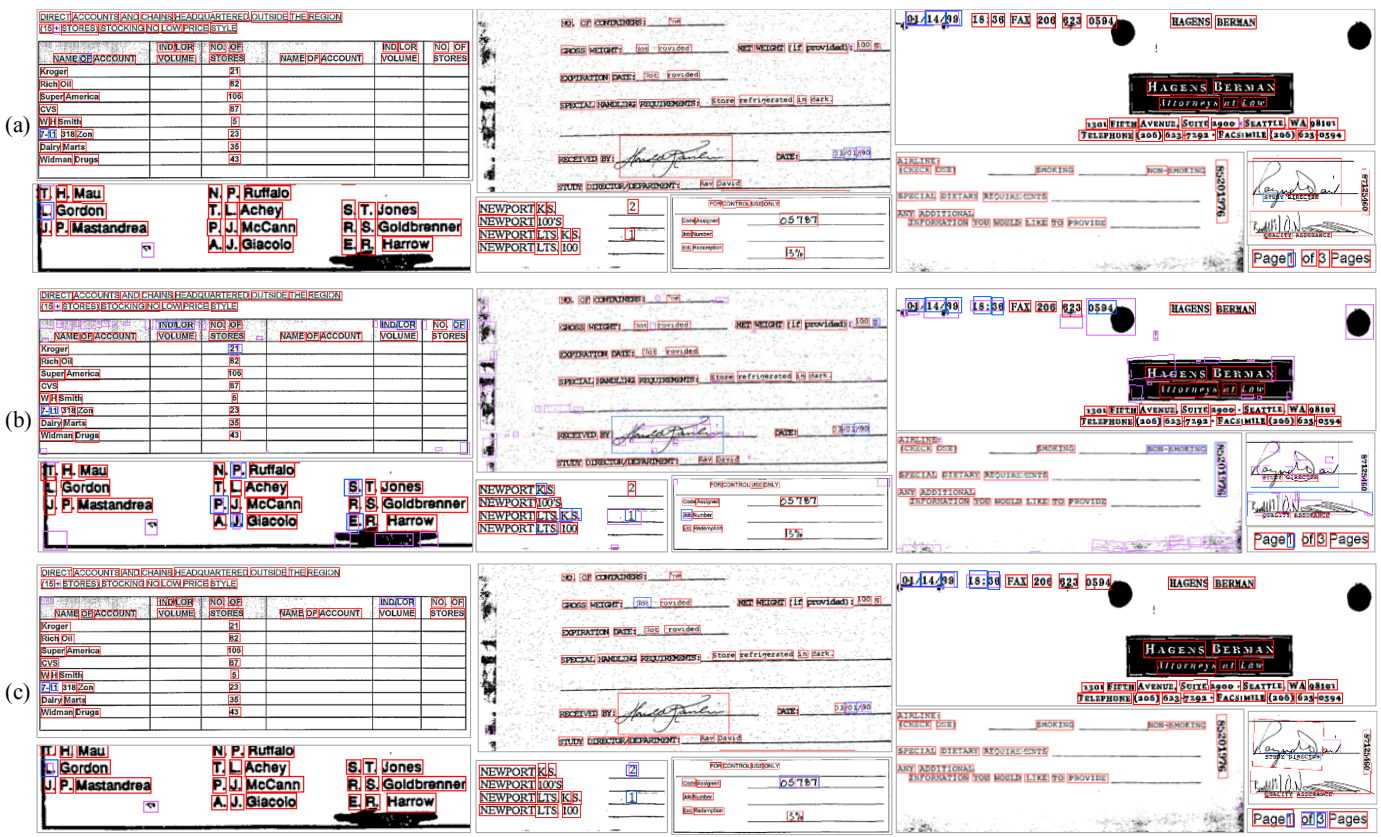

Fig. 4. Text detection results of (a) complete DUET (after training phase-2), (b) DUET after training phase-1 (trained only with synthetic data), and (c) the network without enhancement task (single-task). True detections, false detections, and false rejections are presented in red, magenta, and blue, respectively.

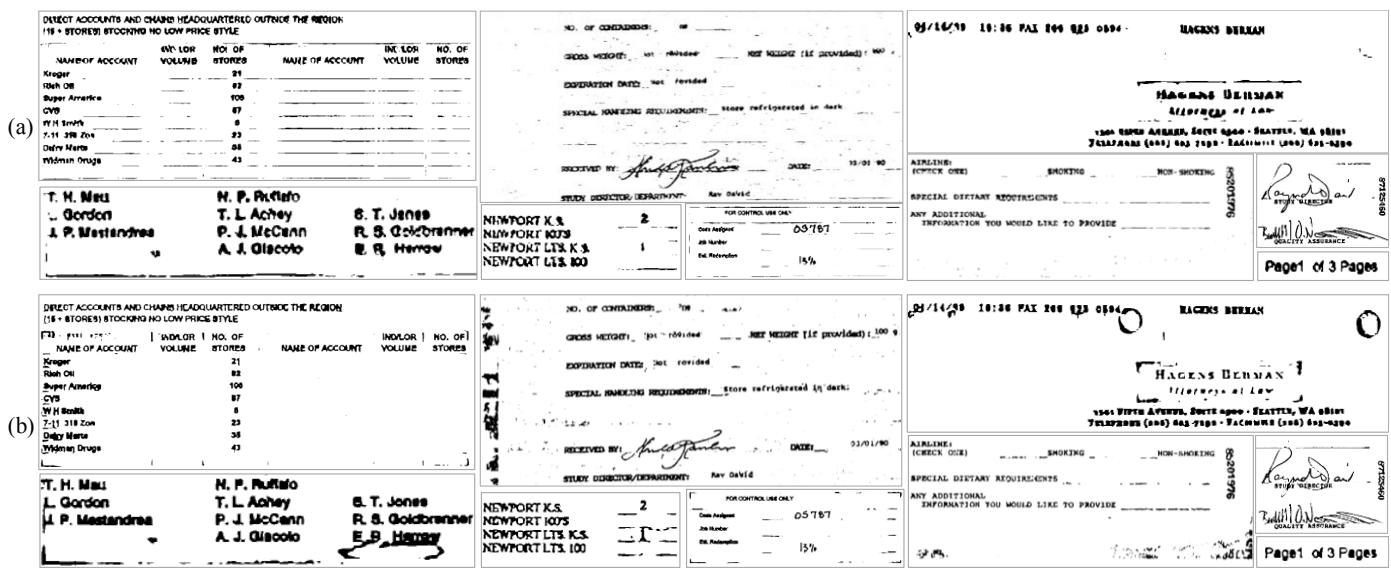

Fig. 5. Text enhancement results of (a) complete DUET (after training phase-2) and (b) DUET after training phase-1 (trained only with synthetic data).

*E. Ablation Test*

We extend our analysis with various ablations. First, the effects of **train data** and **multi-task** are demonstrated together in Table II and examples of detection results in Fig. 4b and c. The network without enhancement branch is also trained as DUET; only synthetic data are trained in the first phase and real data are added in the second phase. Both precision and recall are increased after training real data, however, the increase of precision is more remarkable. Without real train data, many parts of line or background noise are falsely detected as presented in Fig. 4b. Also, the detection performances of the network without the enhancement branch are lower than those with the enhancement branch. We can interpret this result that various features are more robustly trained with the proposed multi-task.

To confirm the effects of proposed **weakly-supervised** learning of real data, enhanced text outputs before training phase-2 are presented in Fig. 5b. Horizontal lines are impressively removed with training of only synthetic data. However, vertical lines, corners of tables, hole punch marks, and background noise are removed remarkably with training data including the real data. These results infer that DUET successfully trained to enhance text without the exact ground truth.

We also observed the detection performances of the two branches of DUET **separately** as shown in Table III; inputs for the detection branch are the original input images and those for the detector after enhancement branch are enhanced outputs. Examples of detection results are visualized in Fig. 6a and b. The detection results with enhanced document images as inputs are lower than those of the original detection. However, a few cases of false detections with the original inputs can be eliminated with text detection of the enhanced version. By adding information of text enhanced output, the detection performance of DUET increases in the precision rate.

TABLE I. TEXT DETECTION RESULTS AND COMPARISONS.

| Method | Precision | Recall | F-score |
|---|---|---|---|
| **Tesseract**[a] [5] | 45.4 | 68.0 | 54.4 |
| **EAST**[a] [12] | 51.6 | 84.0 | 63.9 |
| **Google Vision**[a] (API) | 79.8 | 62.0 | 69.8 |
| **Faster R-CNN**[a] [24] | 70.4 | 84.8 | 76.9 |
| **CRAFT**[b] [13] | 91.2 | 84.2 | 87.6 |
| **CharNet**[b] [14] | **95.1** | 57.4 | 71.6 |
| **DUET** *(proposed)* | 93.1 | **92.2** | **92.6** |

[a.] Results provided from [38]

[b.] Results using the trained models and test codes from the repositories of the original studies

TABLE II. EFFECTS OF TRAIN DATA AND MULTI-TASKING.

| Train data | Synthetic (training phase-1) | | | Synthetic & Real (training phase-2) | | |
|---|---|---|---|---|---|---|
| **Task** | P | R | F | P | R | F |
| **Single** | 77.9 | 85.7 | 81.6 | 92.9 | 90.7 | 91.8 |
| **Multi** | 76.2 | 87.4 | 81.4 | **93.1** | **92.2** | **92.6**[c] |

[c.] Multi-task with synthetic & real training data is same as DUET result in Table I

TABLE III. DETECTION RESULTS OF THE ORIGINAL AND ENHANCED INPUT SEPARATELY.

| Input | Precision | Recall | F-score |
|---|---|---|---|
| **Original** | **92.5** | **92.2** | **92.5** |
| **Enhanced** | 89.0 | 89.8 | 89.4 |

## V. CONCLUSION

In this research, we proposed DUET which is a novel approach of text detection in document images utilizing weakly-supervised text enhancing subtask. While most of the text detection studies focus on that in the wild, we oriented our goal to robust detection of text in documents. To overcome the insufficiency of document image dataset, synthetic data were generated. The efficacy of our synthetic data was suported with the model trained only with synthetic data. Moreover, document enhancing as an auxiliary task also improved DUET's text detection performances. DUET successfully detected densely arranged text segments in real documents with an additional output of the enhanced version of the documents with less noise. We demonstrated our method with a scanned document database to present numerical performances, however, it can be also applied to various types of document images including those captured with cameras. Our work contributes significantly to the applications of document and text image analysis.

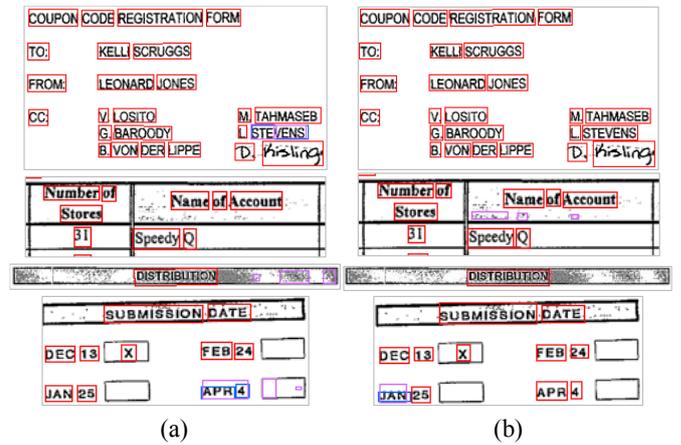

Fig. 6. Text detection results of the (a) detection and (b) enhancement branches of DUET. True detections, false detections, and false rejections are presented in red, magenta, and blue, respectively.